\documentclass[authoryear,preprint,review,12pt]{elsarticle}

\usepackage{graphicx}
\usepackage[ruled,vlined]{algorithm2e}
\usepackage{amssymb}
\journal{Expert System with Applications}

\begin{document}

\begin{frontmatter}


%\title{An Adaptation of Case-Based Reasoning for Real-Time Virtual Agent Decision Making\tnoteref{label1}}
%\tnotetext[label1]{An application on virtual environment for training}
\title{Real-time Retrieval for Case-Based Reasoning in Interactive Multiagent-Based Simulations}
\author{Pierre De Loor\corref{cor1}}
\ead{deloor@enib.fr}
%\ead[url]{www.enib.fr\slash $\sim$deloor}
\author{Romain B\'enard}
\ead{benard@enib.fr}
%\ead[url]{www.enib.fr\slash $\sim$benard}
\author{Pierre Chevaillier}
\ead{chevaillier@enib.fr}
%\ead[url]{www.enib.fr\slash $\sim$chevaillier}

%\fntext[label2]{ test label12}
\cortext[cor1]{Corresponding author. Tel: +33(0)2 98 05 89 60;
fax: +33(0)2 98 05 89 79. \\
Postal Address: Centre Europ\'een de R\'ealit\'e Virtuelle,
26 rue Claude Chappe, 29 280, Plouzan\'e,
France.\\
}
%\address{Adress\fnref{label3}}
\address{UEB - ENIB - LISyC, CERV, Brest, 29200, France \\}

%%\title{}


%%\author{}

%%\address{}

%\section{Corresponding Author}*
%Mr Pierre De Loor\\
%Centre Europ\'en de R\'ealit\'e Virtuelle \\
%26, rue Claude Chappe, 29 280, Plouzan\'e \\
%France \\
%
%\noindent
%t\'el : +33(0)2 98 05 89 60 \\
%fax : +33(0)2 98 05 89 79 \\

\begin{abstract}
The aim of this paper is to present the principles and results of case-based reasoning adaptated to real-time interactive simulations, more precisely concerning retrieval mechanisms. The article begins by introducing the constraints involved in interactive multiagent-based simulations. The second section presents a framework stemming from case-based reasoning by autonomous agents. Each agent uses a case base of local situations and, from this base, it can choose an action in order to interact with other autonomous agents or users' avatars. We illustrate this framework with an example dedicated to the study of dynamic situations in football. We then go on to address the difficulties of conducting such simulations in real-time and propose a model of case and of case base. Using generic agents and adequate case base structure associated with a dedicated recall algorithm, we improve retrieval performance under time pressure compared to classic CBR techniques. We present some results relating to the performance of this solution. The article concludes by outlining future development of our project.
\end{abstract}


\end{frontmatter}


\section{Introduction}
\label{intro}
Videogame technologies have recently begun to be used for the purposes of scientific simulation and visualization (\cite{ferey-2008}), industrial and military training (\cite{gonzalez-1998,Buche04b}), and finally medical and health training and education (\cite{volbracht-1998,Bideau-2003}). Within these simulations, users can interact with autonomous agents and/or human avatars of team members (\cite{Raybourn_iwc07}). %Various combinations of multiple human or virtual trainees may be brought together for practice exercises. 

Unlike video games, these simulations tend not to focus on the quality of graphical representations or animation which are not always necessary for optimizing understanding of these situations (\cite{metoyer-2000}). The most important point is to ensure variability and sponteneity within the simulation. The present paper adresses this issue in dynamic and collaborative situations. Unlike procedural activities, dynamic and collaborative situations cannot easily be defined by sequences of rules as there are an infinite number of possible situations. These situations result from local interaction beetwen participants unaware of the overall situation. It is therefore possible to simulate such dynamics using autonomous agents interacting with one or more users. In this case, decision-making is a rapid process largely influenced by context, and therefore partial perception, time limitations, high stakes, uncertainty, unclear goals, and organizational constraints (\cite{jonsson03,kofod-petersen05:_contex}). Consequently, the outcomes of agents' actions are unpredictable but can be qualified as more or less believable than real-life experience. Moreover, the objective is to simulate adaptive behaviors capable of reacting to many different situations with some variability. 

Credibility depends on psychological and subjective considerations (\cite{loyall-2004}) and is difficult to quantify. Systematic approaches, such as defining an explicit set of rules (\cite{laird-2000}), or automatically learning rules (\cite{sanza-1999}), therefore conflict with believability. Even if these lastest methods are used to define behaviors in simulating collaborative and dynamic situations (\cite{ros-2006}), they are based on the optimization of ``simple'' criteria (for example, an agent's score, or time taken to complete a task). Consequently, the resulting behavior is efficient, but unnatural and unsuitable for human learning.

Another approach is available to interactively construct dynamic and collaborative situations: the use of case-based reasoning (CBR) (\cite{aamodt-1994}) in association with context modeling (\cite{gonzalez-1998,brezillon-1999}, \cite{benard06c}). Case-based reasoning stems from analogous reasoning (\cite{Kolodner1993,Riesbeck1989,Eremeev2006}), which is particularly relevant for addressing decision-making in dynamic and collaborative situations (\cite{bossard-2006}). Context relies on all the elements perceived at any one time by a given agent which might influence its decision-making. This concept arises from ecological psychology (\cite{gibson-1958}) and is strongly linked with naturalistic decision making (\cite{klein-2008}). 
This article mainly addresses the principal difficulty faced when using CBR in this way: maintaining performance in real-time. The time needed to retrieve a case increases with the size of the base multiplied by the number of autonomous agents. For real-time purposes, it is unacceptable for the time taken to make a decision to be linearly dependent on the size of the base (time taken to scan the base), as it is subject to great variation. Depending on the domain in which CBR is applied, the size of the base may increase with experience, or by means of machine learning algorithms executed during experimental sessions etc. 
Moreover, the term ``dynamic situation'' implies that, at any given time, agents must be able to carry out an action even if it is not the best one. Nevertheless, it is important to be aware that, even when these decisions may be inappropriate, they are the result of heuristics and are not merely random. Experts also claim that perceptions guide actions and that not all perceptions are equal, but rather they depend on their implication in the decision (\cite{klein-2008}). It is therefore important to highlight the fact that incorrect or incomplete perceptions may lead to inappropriate actions. Such approximate perception is attributed to a lack of time available to perceive. 

These principles can be implemented using the architecture presented here in this paper. This architecture will be able to model 1) that some perceptions are more relevant than others in making decisions (under time pressure, agents will focus on these perceptions first), and 2) that the shorter the time, the worse the perception, and therefore decisions made due to that perception, will be.

This article is divided into three main parts: section \ref{CBR} describes CBR and a context model associated to each case. An application, \textit{CoPeFoot}, is used to illustrate this proposal. Section \ref{RTCBR} addresses a real-time adaptation of case retrieval. Section \ref{result} shows how this proposition improves recall results and system precision under real-time constraints. These results are also discussed in preparation for the conclusion in section \ref{conclusion}.

\section{CBR for Decision-Making in Virtual Agents}
\label{CBR}
\begin{figure}[!h]
\centering
\includegraphics[width=.7\columnwidth]{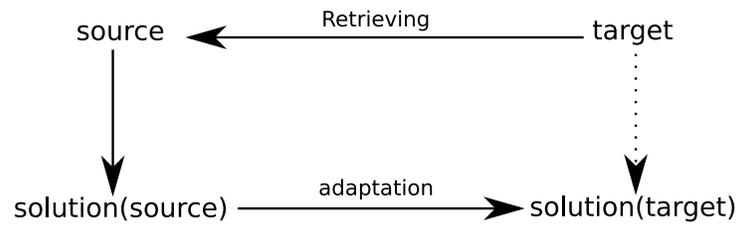}	
\caption{The Principles of CBR}
\label{fig:cbrPrinciple}
\end{figure}

Case-based reasoning stems from analogous reasoning which states that each situation encountered can be associated with another similar well-known and appropriately-resolved situation. The difficulty is in defining how to associate situations in order to choose the most relevant, and to adapt one situation to fit another. The principles of case-based reasoning are summarized in figure \ref{fig:cbrPrinciple}. When the expert system encounters a problem \textit{(case target)} it searches for a similar case in its base \textit{(case source)} which is associated with a solution \textit{(solution(source))}. It then adapts either \textit{(solution(source))} or the resolution derived from case source to \textit{(solution(source))}, in order to define the solution \textit{(solution(target))}. The main advantage is that it is unnecessary to detail an exhaustive resolution mechanism which can become so complex that it is in fact unknown. The adaptation step concerns either the resolving procedure or the solution directly (\cite{Lieber-2007,Cordier2006}).

\begin{figure}
\centering
\includegraphics[width=.8\columnwidth]{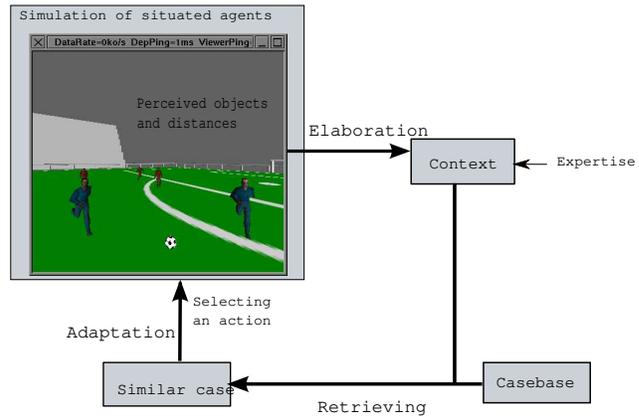}	
\caption{Case-based reasoning within simulation of interactive dynamic situations}
\label{fig:refCoPeFoot}
\end{figure}

An application of CBR to decision-making in autonomous agents in interactive simulations is illustrated in figure \ref{fig:refCoPeFoot}. Each autonomous agent uses CBR to choose its subsequent decisions within the simulation.

The \textit{context box} is the process of abstraction which extracts semantic information from features perceived in the simulated world. More precisely, whereas simulations produce low-level information like changes in the positions of objects, the \textit{context box} gives information such as the qualitative distances between agents (agent \textit{a} is far from agent \textit{b}), or more domain dependent information (see examples in the following section). Each autonomous agent has its own context depending on its position in the virtual world. In CBR, this step is known as elaboration, during which all of the relevant context elements are defined by experts. This context is compared with elements from the case base in order to select one case (``recall step''). Finally, using semantic information, the case is adapted to the current situation and autonomous agents can act within the virtual world (``adaptation step''). Both the elaboration and adaptation steps are part of psychological research linked to the field in which the CBR is implemented.

\subsection{Application}
\begin{figure}
\centering
\includegraphics[width=.5\columnwidth]{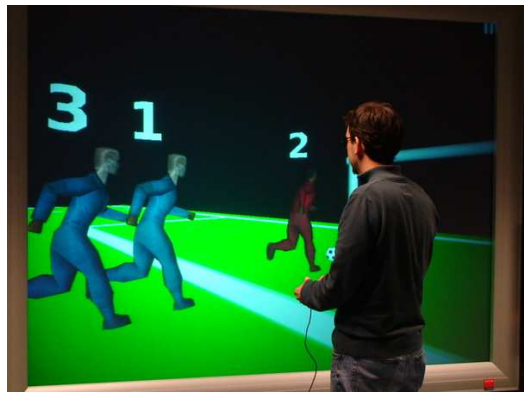}	
\caption{A user interacting with agents in the CoPeFoot simulator}
\label{jeu}
\end{figure}

The theoretical proposition was implemented in the \textit{CoPeFoot} simulation tool for studying collaborative and dynamic situations in sport (\cite{bossard-2006}). This application will be used to illustrate each step of the theoritical model. The pratical uses of \textit{CoPeFoot} are described in (\cite{bossard-2006}). It is designed to be used for training sports coaches and referees. Both the starting conditions and the exercices can be configured in order to immerse a real player in a 3D scene with autonomous players (by means of stereo vision glasses). Users can also study the situations from different points of view by watching the recordings of their movements. Figure \ref{jeu} depicts a user interacting with \textit{CoPeFoot} in an immersive room. 

\subsection{Context Model}
Although context is domain dependent, it is possible to formalize its data structure as follows: a context $Ctx$ is a set of predicates. Each predicate stands for one possible perception of an agent, and is domain-specific. 
Examples of such perceptions for football are the fact that a player is marked (followed by an opponent) or that a team-mate is asking for the ball (asking for the ball is a possible action for any player). 

A predicate $pr_x$ is a triplet $\{n_x,{\cal{S}}_x,c_x\}$ where:
\begin{itemize}
	\item $n_x$ is the name of the perception (for example $distance$ to express the perception of a specific distance).
	\item ${\cal{S}}_x$ is a set of variables. Each variable $v \in {\cal{S}}_x$ has a type $type(v)$ inherited from a generic type \texttt{DomainObject} or an \texttt{Agent} type. This type is interesting because, in dynamic situations, some cases are only different because the identity of some of the agents differ, whereas the other context elements remain the same. In order to reduce the case base size and improve adaptation, it is therefore possible to formalize a \textit{generic case} representing all the cases with the same context except the identity of the agents (see section \ref{genericCase} for details). 
	\item $c_x$ is the variable which expresses the \textit{intended} value of the predicate. It is used when transforming the case base into a hierachical tree (see section \ref{RTCBR}). Two types are possible:
	\begin{itemize}
		\item $Boolean$: expresses the success of a Boolean test linked with the purpose of the predicate.
		\item $Qualitative Value$: an abstraction of a number (e.g. a distance beetwen 8 and 20 meters is equivalent to the qualitative value ``far'') and enables the aggregation of a number of values into one when considered by experts to be similar enough.
	\end{itemize} 
\end{itemize}

Variables are instantiated in real-time by the simulator using a prolog-like mechanism when the CBR enquires about the validity of predicates during recall. %

\subsection{Application to \textit{CoPeFoot}}
Table 1 shows the predicates engaged at a specific moment in the \textit{CoPeFoot} simulator. Some predicates like $distance$ or $relativePosition$ or $orientation$ are very general and applied to a type $Physical Object$ which represents all possible objects perceived in the 3D scene. Other predicates are more specific to football ($hasBall$, $isMarked$, $markedBy$, $callForBall$, $callForSupport$, $partner$, $isInAttack$, $ratio$, $lastAction$). They use $Agent$ variables which represent football players. Other football-specific predicate types are $Team$, $Action$, $Ball$, $Goal$ or $Field$ (these last three  types inherit from $Physical Object$). 
For example, when CBR asks the simulator about the validity of one predicate $\{\{isMarked,\{X\},true\}\}$ which means \textit{is a marked football player currently perceived?}, the simulator answers \texttt{true} with an instanciation: \texttt{X=Agent.3} (in this case, \texttt{Agent.3} is an instance of the Agent class of the simulation which is perceived by the Agent decision-maker and marked by an opponent). If no players are marked, the simulator answers \texttt{false}.  
 
%\begin{table}
%\begin{tabular}{|c|c|c|}
%\hline
%$n$ & \cal{S} & $c$ \\
%& (with their types) & (with its type) \\
%\hline
%\hline
%distance & \{O1 (Physical object), O2 (Physical object) \}& D (Qualitative value)\\ 
%\hline
%relativePosition & \{O1 (Physical object), O2 (Physical object) \}& O (Qualitative value)\\ 
%\hline
%orientation & \{ O1 (Physical object) \}& O (Qualitative value)\\ 
%\hline
%hasBall & \{P1 (Agent) \}& V (Boolean) \\
%\hline
%isMarked & \{P1 (Agent) \}& V (Boolean) \\
%\hline
%markedBy & \{P1 (Agent), P2 (Agent) \}& V (Boolean) \\
%\hline
%callForBall & \{P1 (Agent) \}& V (Boolean) \\
%\hline
%callForSupport & \{P1 (Agent) \}& V (Boolean) \\
%\hline
%partner & \{P1 (Agent) \}& V (Boolean) \\
%\hline
%isInAttack & \{P1 (Agent) \}& V (Boolean) \\
%\hline
%ratio & \{DO1 (Team) \}& N (Qualitative value) \\
%\hline
%lastAction & \{DO1 (Action) \}& B (Boolean) \\
%\hline
%
%\end{tabular}
%\label{tab:predicateContext}
%\caption{Predicates expressing the context of an agent in \textit{CoPeFoot}. For example, the first predicate expresses a distance between two objects and is formalized by $pr_x = \{\{distance,(A,B),L\}\}$. When $type(A)=type(B)=Physical Object$, $type(L)=Qualitative Value$ (for a distance, qualitative values of this variable are \textit{\{close, far, long\})}. 
%}
%\end{table}

\subsection{Generic Case Model}
\label{genericCase}
In the following section, we will outline the model of a \textit{generic case} using \textit{generic agents} and an \textit{instanciated case} which corresponds to a concrete situation with \textit{instanciated agents}. The case base, representing source cases, is made up of a set of \textit{generic cases} whereas \textit{instanciated cases} are elaborated from an ``instantaneous'' moment of the situation in the simulation; they correspond to target cases.

\label{caseModel}
A generic case $c \in CaseBase$ is a triplet $c = \{\cal{P},\cal{W}\}$, where: 

\begin{itemize}
 \item  $\cal{P}$ is a set of the agent's perceptions $p_i$. A perception $p_i$ is a triplet $\{n_i,{\cal{V}}_i,vc_i\}$ where:
 \begin{itemize} 
 \item $n_i$ is the name of a predicate $pr_i \in Ctx$
 \item ${\cal{V}}_i$ is the set of values of the variables $i \in pr_i$. 
 \item $vc_i$ is the value of $c_i$.
 \end{itemize}
 
In generic cases, $Agent$ variables are not instances of the simulation (like \texttt{Agent.1} or \texttt{Agent.2}) but are rather \textit{generic agents} representing every possible agent in the simulation. Predicates can thus be linked together in the case. For example, in \textit{CoPeFoot}, a case can specify that the marked \textit{generic agent \texttt{A}} is the same as the agent in possession of the ball, but that another \textit{generic agent \texttt{B}} is far from the first (in this case ${\cal{P}}=\{\{hasBall,\{A\},true\},\{isMarked,\{A\},true\},\{distance,\{A,B\},far\}\}$). It is worth noting that there is one exception: the agent $me$, the decision maker, is not instanciated by a \textit{generic agent}. Indeed, to obtain consistent decision-making, it is important to distinguish this agent from all the others, even in a \textit{generic case}.

This case base is heterogenous, i.e. the number of perceptions relative to a case and the type of perceptions differ for each case. Indeed, it depends on the situation/orientation/position of the player who will not always perceive the same elements.  

\item $\cal{W}$ is a set of predicate relevance weights: $\{w_{{p_1}{c_k}},w_{{p_2}{c_k}},w_{{p_i}{c_k}}...,w_{{p_m}{c_k}}\}$. Where:
\begin{itemize}
\item $p_i$ is a perception of the generic case $c$
\item $c_k$ is a case from the case base
\end{itemize}
 Hence, a weight $w_{{p_i}{c_k}}$  is a real value, representative of the relevance of the perception $p_i$ for the case $c_k$. This relevance must be defined by experts. The classical similarity function between two cases $c_k$ and $c_l$ is adapted for cases with non-constant numbers of parameters, and is described by equation \ref{eq:similarity} in section \ref{similarity}.

\end{itemize}

Figures \ref{fig:XMLCtxt} and \ref{fig:XMLCases}, respectively give a simple example from \textit{CoPeFoot} written in XML, for the context ``$\{\{hasball,\{Y1\},Y2\},\{partner,\{X1\},X2\},$ $\{distance,\{Z1,Z2\},Z3\}\}$'' and of one generic case relying on this context.

\subsection{Instanciated Case Model}
An {\it instanciated case} arises from a {\it generic case} with \textit{generic agents} substituted by instances of the simulation (noted \texttt{Agent.1}, \texttt{Agent.2} ...etc). Generally, they constitute target cases.

An example from \textit{CoPeFoot} of a case which matches the example in figure \ref{fig:XMLCases} is $\{case1=\{\{\{hasBall,\{me\},false\}, \{partner,\{Agent.1\},true\},\\ \{distance,\{ball,Agent.1\},long\}\}\}\}$. 

\subsection{Similarity}
\label{similarity}
From a source case $c_s=\{{\cal{P}}_s,{\cal{W}}_s\}$ and a target case represented by a perception set ${\cal{P}}_t$, we define $idP_{st}$, the set of common perceptions of ${\cal{P}}_s$ and ${\cal{P}}_t$ when players of ${\cal{P}}_s$ are instanciated with players of ${\cal{P}}_t$ : $idP_{st}={\cal{P}}_s \cap  {\cal{P}}_t$.
The similarity value $sim_{st}$ beetwen the source case $s$ and the target case $t$ is defined by equation \ref{eq:similarity}.

\begin{eqnarray}
   sim_{st}=\frac{{\sum_{i \in {idP_{st}}} w_{is}}}{{\sum_{i \in {{\cal{P}}_s}} w_{is}}} \times (1-\alpha \frac{|{\cal{P}}_t|-|idP_{st}|}{|{\cal{P}}_t|}) \label{eq:similarity} 
\end{eqnarray}

This equation takes into account the base's heterogeneous nature. It is therefore possible to compare the similarity between two source cases and the target case even if the sources and the target differ in their numbers of perceptions. In such a case, the nearer the number of identical perceptions $|idP_{st}|$ is to the case size $|{\cal{P}}_t|$, the greater the similarity. $\alpha$ is a weighting parameter with a value between 0 and 1. Its influence on similarity, recall and precision is explained in section \ref{result}. 

\section{Anytime CBR}
\label{RTCBR}
\subsection{Search-Tree for Real-Time Selection}
Theoretically, recall implies comparing an instantaneous target case with all the generic source cases in the base using the similarity function, in order to select the most similar source case.
In an interactive simulation, the agent needs to make a decision at a precise moment under time pressure. This is dependent on processes like updating (refreshing) the 3D scene, the other agents' decision-making (for example, 22 agents within the context of football), and interaction with users. Moreover, the base size could increase as the application improves both in terms of function and expertise. 

Base scanning must be optimized for both quantity (reducing the number of tests) and  quality (addressing first the relevant perception to make an \textit{adequate} decision).
Here we shall consider a brief summary of this mechanism. It is based on the following facts: 
\begin{itemize}
\item Some variables are shared by different predicates. In these cases it is possible to evaluate each case in parallel. 
\item It is possible, with the help of experts, to define a global order of priority between the perceptions (predicates) taken into account by a player when identifying a case. This order improves both the retrieval speed and the adaptability of selected cases. Less adaptable and domain-specific perceptions are found at the top of the tree. Let us take an example from football. One crucial piece of information for making appropriate decisions is to know if the player in possession of the ball is in our team or not. Furthermore, it is more difficult for another player to change the player who is in possession of the ball than it is to reduce the distance between himself and that player. Consequently, the predicate representing the perception of the player in possession of the ball is positioned higher in the tree than the predicate relative to the distance from this player.
\end{itemize}
Hence, perceptions are grouped together in a tree which re-organizes the case base. 
Figure \ref{fig:treeExample} shows an example of one such tree relative to: 
\begin{enumerate}
\item the context $Ctx$ of figure \ref{fig:XMLCtxt}, 
\item three generic cases (for simplicity the set of pertinence weights ${\cal{W}}_i$ are not defined here):
	\begin{itemize}
	\item \textit{case1=\{\{\{hasBall,\{me\},false\},\{partner,\{A\},true\},\{distance,\{ball,A\},long\}\}, ${\cal{W}}_1$\}}
	\item \textit{case2=\{\{\{hasBall,\{me\},true\},\{partner,\{B\},true\}\}, ${\cal{W}}_2$\}}
	\item \textit{case3=\{\{\{hasBall,\{me\},false\},\{partner,\{A\},false\},\{distance,\{ball,B\},long\}\}, ${\cal{W}}_3$\}}
	\end{itemize}
\item an order of priority : ${hasball>partner>distance}$
\end{enumerate}

Each node represents a predicate $pr_i \in Ctx$, with each variable of ${\cal{S}}_i$ being instanciated according to the value corresponding to each individual case. Each arrow is an instantiation of the $c_i \in pr_i$ which contributes to its validation according to case base. Some nodes are shared by different cases but each case corresponds to one branch. It must be noted that this example is simplified in order to illustrate the principle; typically, trees such as these contain about one hundred nodes. 

Technically, algorithm \ref{algo} constructs such a tree from the case base and the priority order beetween predicates. This tree is built only once, at the beginning of the simulation, and is then used for real-time case retrieval. 

\begin{algorithm}[-htbp]
	\SetLine
	\KwData{
		generic cases $c_i=\{{\cal{P}}_i,{\cal{W}}_i\} \in CaseBase$\;
		context $Ctx=\{pr_1,pr_2,..pr_m\}$}
		priority order of $Ctx$
		
	\KwResult{Hierarchical tree of generic cases}
	\Begin{
		Create an empty node \textit{root}\;
		$root \gets CurrentNode$\;
		\ForAll{$c_i=\{{\cal{P}}_i,{\cal{W}}_i\} \in CaseBase$}{
			\ForAll{$p_j=\{pr_j,{\cal{V}}_j,vc_j\} \in {\cal{P}}_i$ in the priority order}{
			 \If{($CurrentNode$ is labelled with $\{pr_j,{\cal{V}}_j\}$)}{
			 	\If{(exists an arc starting from $CurrentNode$, ending with a node \texttt{n} and labelled with the test $[c_j=vc_j]$)}{
			 		$\texttt{n} \gets CurrentNode$\;
			 		}
			 	\Else{
			 		Create an arc from $CurrentNode$ to a new empty node $n$ labelled with the test $[c_j=vc_j]$\;
					$n \gets CurrentNode$} 
				}
				\Else{
				
				Labelling $CurrentNode$ with $\{pr_j,{\cal{V}}_j\}$\;  
				Create an empty node $n$ and an arc from $CurrentNode$ to $n$ labelled with the test $[c_j=vc_j]$\;
				$n \gets CurrentNode$\;
				}
			}
		Labelling $CurrentNode$ ( a leaf of the tree) with the name $c_i$\;
		$root \gets CurrentNode$\;		
		}
	}
\caption{Building the generic cases tree: note that ($\forall p_i\in {\cal{P}}_i,\exists pr_j \in Ctx$ such that $p_i=\{n_i,c_i,vc_i\}$ and $pr_j=\{n_j,{\cal{S}}_j,c_j\} $ and $i=j$)\label{algo}}
\end{algorithm}

\subsection{Evaluating Similarity in Real-Time}
The tree is scanned widthwise during the simulation to identify cases similar to the current situation. This scanning is interrupted by the simulator when it is the turn of the corresponding autonomous agent to make a decision. Consequently, the most similar case in the base must be available at any time. In order to do so,  requests are sent to the context box, starting from the root (see figure \ref{fig:refCoPeFoot}). In response, the values of variables validating the perception of each scanned node $i$ (${\cal{S}}_i$ and $c_i$) are transmitted, if they exist. If they do not exist, the variables remain un-instanciated in the tree. The longer the delay beetwen two interruptions, the higher the number of scanned nodes and instanciated variables. If an arc is invalidated by the value of its corresponding tested variable ($c_i$), the branch is pruned and the similarities of corresponding cases are no longer evaluated. \\

For one source case $c_s=\{{\cal{P}}_s,{\cal{W}}_s\}$, the set of nodes which are along the branch from the root to the final node corresponds to ${\cal{P}}_s$. In section \ref{similarity}, we defined the similarity function beetwen such a case and a target case represented by a perception set ${\cal{P}}_t$ (equation \ref{eq:similarity}). It is possible to adapt this definition for real-time similarity evaluation in the following manner: 
${\cal{P}}_s(t) \subseteq {\cal{P}}_s$ can be defined as is the set of perceptions of $c_s$ which are already scanned (starting from the root) at time $t$. Similarly, it is possible to define $idP_{st}(t)= {\cal{P}}_s(t) \cap  {\cal{P}}_t$.
This definition implies that:\\

\noindent
\begin{eqnarray}
\lim_{t \to 0} idP_{st}(t) = \emptyset 
\end{eqnarray}
\begin{eqnarray}
\lim_{t \to\infty} idP_{st}(t) = idP_{st}
\end{eqnarray}

The real-time similarity function beetween $c_s=\{{\cal{P}}_s,{\cal{W}}_s\}$ and a target case defined by a set ${\cal{P}}_t$ is given in equation \ref{eq:rt_similarity}.

\begin{eqnarray}
 sim_{st}(t)=\frac{{\sum_{i \in {idP_{st}(t)}} w_{is}}}{{\sum_{i \in {{\cal{P}}_s}} w_{is}}} \times (1-\alpha \frac{|{\cal{P}}_t|-|idP_{st}(t)|}{|{\cal{P}}_t|}) \label{eq:rt_similarity} 
\end{eqnarray}

\section{Results}
The aim of the following experiment is to illustrate the performance of this proposition as compared to a classical approach in terms of real-time performance for recall and precision criteria. The first part evaluates the improvement in terms of memory size. The second part shows the influence of the $\alpha$ parameter on precision and recall. Finally, we adress variations in recall and precision relative to time taken for the algorithm to retrieve similar source cases. To obtain these results, we used two different types of bases representing the same cases. The first, \textit{linear Base}, was a standard base, composed of a list of cases that were an enumeration of perceptions such as those shown in figure \ref{fig:XMLCases}. The second base, \textit{tree Base}, was a tree obtained with the algorithm \ref{algo} applied to the \textit{linear Base}.
  
\label{result}
\subsection{Case Base Size}
\label{CBSize}
The first results deal with the size of the case base and, more precisely, memory space gained due to the base's tree structure. To obtain these curves, an increasing number of cases are introduced in the two bases (\textit{tree Base} and \textit{linear Base}). Cases in \textit{CoPeFoot} occur in the following manner: from an empty case base, users control avatars, and every time they act (press a button), the system records the corresponding case and stores it in the base. Figure \ref{fig:courbesMemoire} shows the evolution of the number of perceptions stored in bases during acquisition. The dotted lines represent \textit{linear Base} and the continuous line \textit{tree Base}.

\subsection{Impact of $\alpha$ on Recall and Precision}
The results introduced in this section are identical for \textit{linear Base} and \textit{tree Base}. They indicate the ways in which it is possible to favor recall relative to precision. In equation \ref{eq:similarity}, similarity depends on a parameter, $\alpha$. For this measurement, experts define a set of cases which are similar to a target case, $t$. This pre-defined set is called $C_1$. The set of cases found through case retrieval is called $C_2$. To carry out this set, similarity function \ref{eq:similarity} is combined with an acceptance threshold. Similarity functions use relevant weights $\cal{W}$ which are also defined by an expert. %They also define the base's weights W (see section \ref{})
The acceptance threshold $th_t$ corresponds to the least significant similarity beetwen each case in $C_2$ and the target $t$ : $th_t = \min \{sim_{st}, s \in C_2\}$.

As in (\cite{kumar-2009}), recall and precision are defined as:  
\begin{eqnarray}
recall=\frac{N_{correct}}{N_{correct}+N_{false}}
\end{eqnarray}
\begin{eqnarray}
precision=\frac{N_{correct}}{N_{total}}
\end{eqnarray}
with:
\begin{itemize}
\item $N_{total}=N_{correct}+N_{missed}$
\item $N_{correct} = |C_1 \cap C_2|$, number of cases correctly found by the algorithm relative to the expert's definitions.
\item $N_{false}=|C_2|-|C_1 \cap C_2|$, number of cases which should not have been considered (expert did not consider them similar to the target)
\item $N_{missed}=|C_1|-|C_1 \cap C_2|$, cases not found by the algorithm.
\end{itemize}

Figure \ref{fig:influence_alpha} shows that depending on the $\alpha$ parameter, we can favor either recall or precision. The nearer $\alpha$ is to 1, the worse the similarities: recall is poor and precision is great. On the contrary, the closer $\alpha$ is to 0, the better the similarities are: recall is optimized and precision is poor. The best results are obtained with $\alpha=0.5$ which is the value used to define the acceptance threshold $th_t$. For the following evaluations, $\alpha$ is set at 0.5.

\subsection{Real-Time Performance}
\label{rtperformances}
In this section, we compare real-time performance between the \textit{linear Base} and the \textit{tree Base} for both recall and precision. The evaluation of both bases is illustred in figure \ref{fig:courbesTR}. In the case of the \textit{linear Base}, the shorter the time, the smaller the number of evaluated cases. This impacts negatively on both recall and precision. To prevent bias linked to case scanning order, cases are chosen randomly and the calculation repeated one hundred times. The curve indicates average precision and recall. With \textit{tree Base}, the order of perceptions is fixed following an analysis by the expert. Similarly to \textit{linear base}, the shorter the time, the less nodes $idP_{st}(t)$ are taken into account. This negatively impacts on similarity but a value can be estimated for every case using equation \ref{eq:rt_similarity}.  

The system took between 2 and 10 ms to evaluate the most similar cases. Simulations were performed with a dual-core processor (3,4 GHz with 2Gb of memory). With more than 10 ms the system has enough time to scan and evaluate the entire \textit{linear Base} and all the nodes of the \textit{treeBase}. In order to enhance the credibility of the simulation, it is important not to leave too much time between two decisions.
For example, in \textit{CoPeFoot}, with 10 autonomous agents, all of the decision-making calculations for the whole team can be acheived in between 20 and 100 ms. For interactive simulation, a frame rate of 25 images per second is generally considered acceptable.

Figure \ref{fig:courbesTR} shows that recall obtained with \textit{linear Base} decreases to 8ms. At this speed, not all of the cases from the base could be evaluated, so precision also begins to decrease. The situation is better for the \textit{tree Base} since precision and recall decrease to 4.2 and 3.4 ms respectively. Nevertheless, when precision and recall values for \textit{tree Base} decrease, they fall sharply until they are poorer than those of the \textit{linear Base}. This can easily be explained since, when processing the tree, if the number of nodes to be processed is so low (falls below a threshold value) that there are not enough significant nodes to be able to calculate the similarities, this affects all of the cases within the base and not only some of them, as is the case in the \textit{linear Base}. %It may be surprising that this decrease comes so rapidly compared to \textit{linear Base}. 
Indeed, within this time (4.2 ms), it is possible to scan 1/10 of the \textit{linear Base}, but this is not the case for \textit{Tree Base}. Nevertheless, section \ref{CBSize} showed that \textit{tree Base} is smaller than \textit{linear Base}. Problems stem from the fact that it takes a long time to browse tree nodes. To illustrate this aspect, figure \ref{fig:courbesProf} shows the same results as figure \ref{fig:courbesTR}. However, this time, the abscissa do not indicate real-time but rather the number of comparisons between the perceptions given to the system. Due to this fact, \textit{treeBase} performance is better than that of \textit{linear Base}.

It is worth noting that in these experiments, the case base contains only 50 cases, to ease the work of the experts defining the $C_1$ set. The base size is much larger when simulating a football game. The bigger the base, the greater the difference between \textit{treeBase} and \textit{linearBase} will be. In this case, the curves in figure \ref{fig:courbesTR} would be much more similar to those in figure \ref{fig:courbesProf}.

\section{Conclusions}
\label{conclusion}
Real-time constraints are a critical problem for CBR use in interactive simulations. However, experts claim that it is an appropriate solution for modeling relevant decision-making in simulations of collaborative and dynamic situations (\cite{Bossard-2007}). The main topic of this article is real-time case retrieval. The proposed solution stems from arborescent case bases, which enable the similarity between a target situation and all the cases in the base to be calculated at any time. The precision of this calculation depends on the time allocated for it but unlike the use of a \textit{linear Base}, there is a proposed value for each case at any given time. Moreover, in accordance with psychological considerations, the longer the time allowed, the better the evaluation of a situation. Some prelimilary tests confirmed the credibility of the resulting behavior in \textit{CoPeFoot} (\cite{bossard-2009}).

Case genericity, obtained by the use of a \textit{generic agent}, means that a general configuration can be adapted to a given \textit{instanciated} target situation coming from the simulator. In order to do so, all \textit{instanciated} players are \textit{unified} with the corresponding \textit{generic agent}. These \textit{instanciated} agents are part of \textit{generic actions}, which are not detailed in this article. Indeed, for the moment, we are testing some further definitions of \textit{generic action} in order to better formalize adaptation, but further coordination with experts is required. We are currently developing an expert module of \textit{CoPeFoot}: \textit{ExPeCoPeFoot}. This new module facilitates the interactions beetwen the simulator and experts. They can refine the case base (weight of $\cal{W}$, set of relevant perceptions) during the course of the simulation.



%\begin{thebibliography}{plain}
\bibliographystyle{elsarticle-harv}
\bibliography{biblio}


%\bibitem{}

%\end{thebibliography}
\newpage

\section{Table}
 
\begin{table}[!h]
\begin{tabular}{|c|c|c|}
\hline
$n$ & \cal{S} & $c$ \\
& (with their types) & (with its type) \\
\hline
\hline
distance & \{O1 (Physical object), O2 (Physical object) \}& D (Qualitative value)\\ 
\hline
relativePosition & \{O1 (Physical object), O2 (Physical object) \}& O (Qualitative value)\\ 
\hline
orientation & \{ O1 (Physical object) \}& O (Qualitative value)\\ 
\hline
hasBall & \{P1 (Agent) \}& V (Boolean) \\
\hline
isMarked & \{P1 (Agent) \}& V (Boolean) \\
\hline
markedBy & \{P1 (Agent), P2 (Agent) \}& V (Boolean) \\
\hline
callForBall & \{P1 (Agent) \}& V (Boolean) \\
\hline
callForSupport & \{P1 (Agent) \}& V (Boolean) \\
\hline
partner & \{P1 (Agent) \}& V (Boolean) \\
\hline
isInAttack & \{P1 (Agent) \}& V (Boolean) \\
\hline
ratio & \{DO1 (Team) \}& N (Qualitative value) \\
\hline
lastAction & \{DO1 (Action) \}& B (Boolean) \\
\hline

\end{tabular}
\label{tab:predicateContext}
\caption{Predicates expressing the context of an agent in \textit{CoPeFoot}. For example, the first predicate expresses a distance between two objects and is formalized by $pr_x = \{\{distance,(A,B),D\}\}$. When $type(A)=type(B)=Physical Object$, $type(D)=Qualitative Value$ (for a distance, qualitative values of this variable are \textit{\{close, far, long\})}. 
}
\end{table}

\newpage

\section{Figure captions}
\subsection*{Caption of figure 1}
The Principles of CBR.
\subsection*{Caption of figure 2}
Case-based reasoning within simulations of interactive dynamic situations.
\subsection*{Caption of figure 3}
A user interacting with agents in the CoPeFoot simulator.
\subsection*{Caption of figure 4}
Example of XML formalization of a small context in \textit{CoPeFoot}: \\ $\{\{hasball,\{Y1\},Y2\},\{partner,\{X1\},X2\},\{distance,\{Z1,Z2\},Z3\}\}$
\subsection*{Caption of figure 5}
XML formalization of a generic case in \textit{CoPeFoot}. \\ $case1=\{\{\{hasBall,\{me\},false\},\{partner,\{A\},true\},\{distance,\{ball,A\},long\}\},$ \\ $\{0,3;0,7;0,45\}\}\}$. The main difference with the context is that the variables are instanciated (with generic agents for agent variables and with constants for all other variables).
\subsection*{Caption of figure 6}
Example of a case base tree.
\subsection*{Caption of figure 7}
Number of perceptions stored in \textit{linearBase} (dotted line) relative to \textit{treeBase} (continuous line) during acquisition of new cases.
\subsection*{Caption of figure 8}
Influence of the $\alpha$ parameter on precision and recall (see text for details).
When $alpha$ changes from 0 to 1 precision decreases and recall increases.
\subsection*{Caption of figure 9}
Precision and recall under time pressure (in milliseconds) for classical and perceptions tree similarities.

\newpage
\section{Figures}

\begin{figure}
\centering
\begin{verbatim}
<ctx>
	<predicate>
	  <name=hasball>
	  <variable name=Y1 type=Agent />
	  <choiceVariable name=Y2 type=Boolean /> 
	 </predicate>
	 <predicate>
	  <name> partner </name>
	  <variable name=X1 type=Agent />
	  <choiceVariable name=X2 type=Boolean />
	 </predicate>
	 <predicate>
	  <name> distance </name>
	  <variable name=Z1 type=PhysicalObject />
	  <variable name=Z2 type=Agent />
	  <choiceVariable name=Z3 type=QualitativeValue value=far />
	 </predicate>
</ctx>
\end{verbatim}	
\caption{Example of XML formalization of a small context in \textit{CoPeFoot}: $\{\{hasball,\{Y1\},Y2\},\{partner,\{X1\},X2\},\{distance,\{Z1,Z2\},Z3\}\}$}
\label{fig:XMLCtxt}
\end{figure}

\begin{figure}
\centering
\begin{verbatim}
<case id=case1>
	 <predicate>
	   <name> hasball </name>
	   <value val="me" type=GenericAgent />
	   <valueChoice val=false type=Boolean /> 
	   <w11 val=0,3 />
	 </predicate>
	 <pred>
	  <name> partner </name>
	  <value val=A type=GenericAgent />
	  <valueChoice val=true type=Boolean />
	  <w21 val=0,7 />
	 </predicate>
	 <predicate>
	 	<name> distance </name>
	 	<value val=ball type=Ball />
	  <value val=A type=GenericAgent />
	  <valueChoice val=long type=QualitativeValue values={close,far,long} />
	  <w31 val=0,45 />
	 </predicate>
</case> 
\end{verbatim}	
\caption{XML formalization of a generic case in \textit{CoPeFoot}. $case1=\{\{\{hasBall,\{me\},false\},\{partner,\{A\},true\},\{distance,\{ball,A\},long\}\}, \{0,3;0,7;0,45\}\}\}$. The main difference with the context is that the variables are instanciated (with generic agents for agent variables and with constants for all other variables)}.
\label{fig:XMLCases}
\end{figure}

\begin{figure}
\centering
\includegraphics[width=0.8\textwidth]{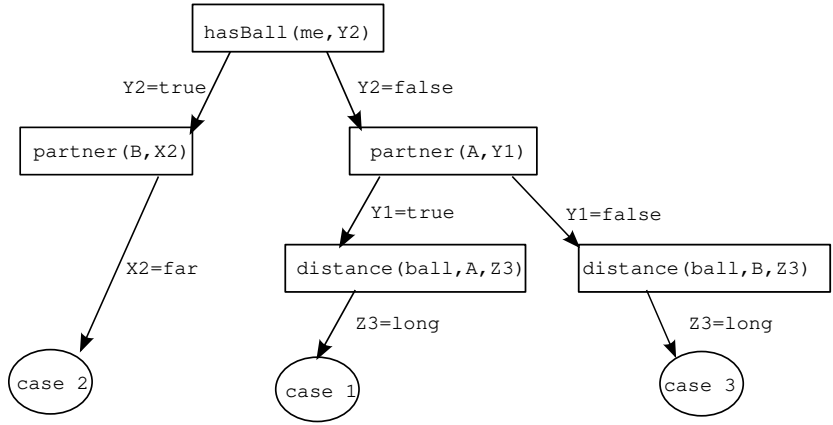}	
\caption{Example of a case base tree}
\label{fig:treeExample}
\end{figure}

\begin{figure}
\centering
\includegraphics[width=.5\columnwidth, angle=270]{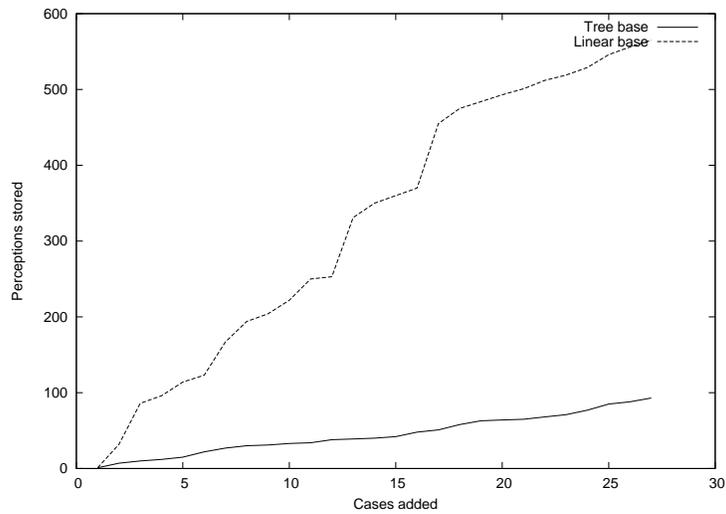}	
\caption{Number of perceptions stored in \textit{linearBase} (dotted line) relative to \textit{treeBase} (continuous line) during acquisition of new cases.
\label{fig:courbesMemoire}}
\end{figure}

\begin{figure}
\centering
\includegraphics[width=.7\columnwidth, angle=270]{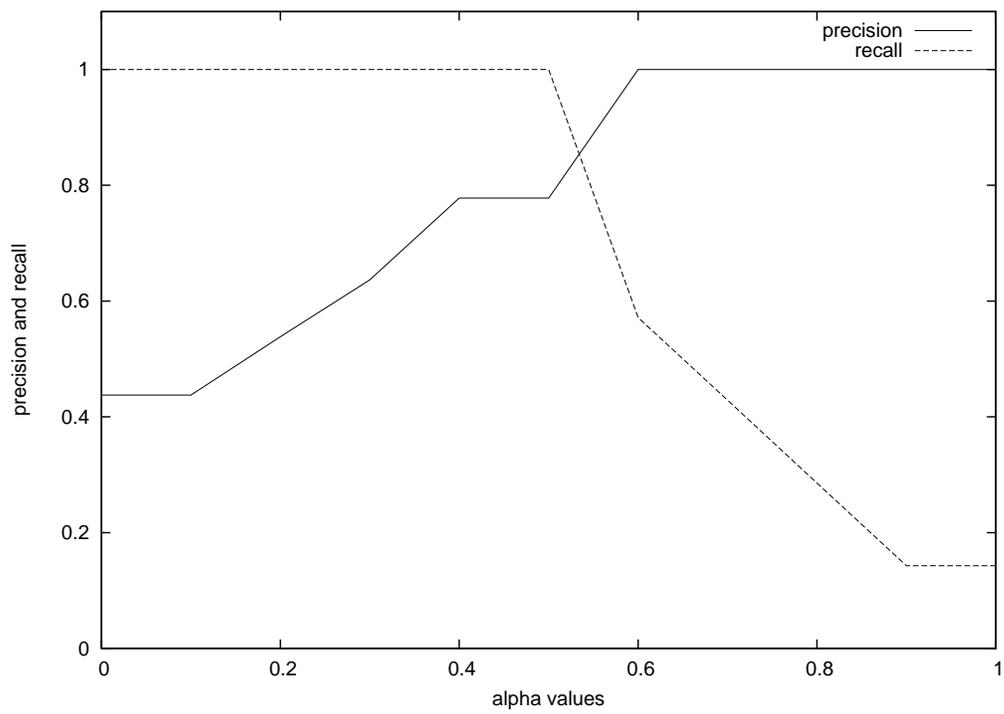}	
\caption{Influence of the $\alpha$ parameter on precision and recall (see text for details).%We fix an $alpha_{threesold}$ to define a threesold of accessibility of the target cases from experts. 
When $\alpha$ changes from 0 to 1 precision decreases and recall increases.}
\label{fig:influence_alpha}
\end{figure}

\begin{figure}
\centering
\includegraphics[width=.7\columnwidth, angle=270]{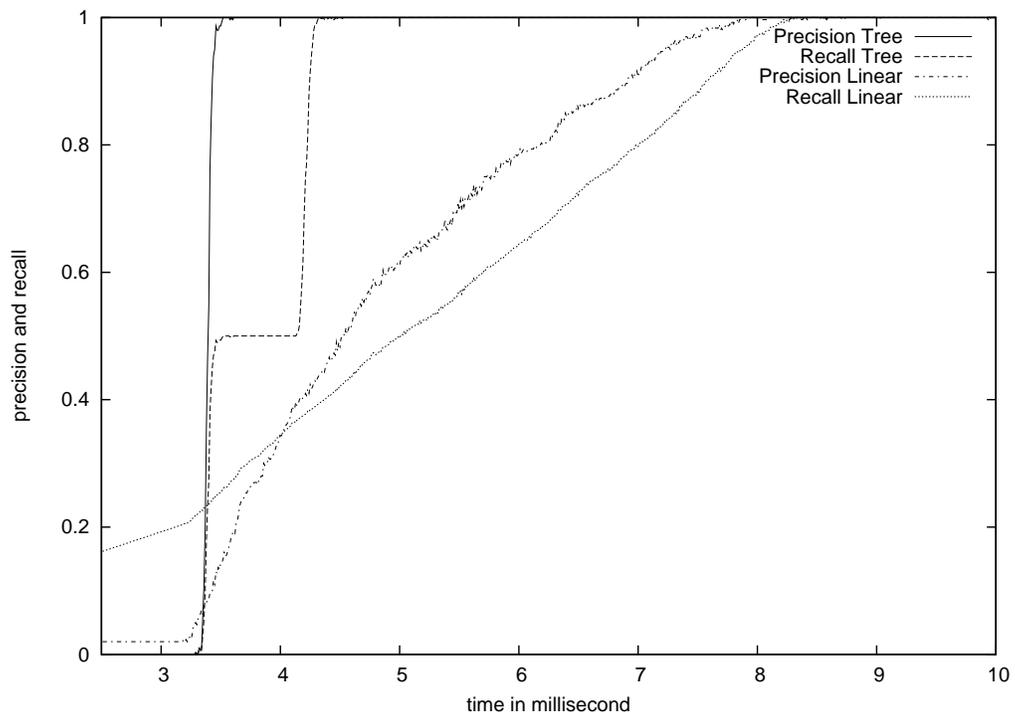}	
\caption{Precision and recall under time pressure (in milliseconds) for classical and perceptions tree similarities}
\label{fig:courbesTR}
\end{figure}

\begin{figure}
\centering
\includegraphics[width=.7\columnwidth, angle=270]{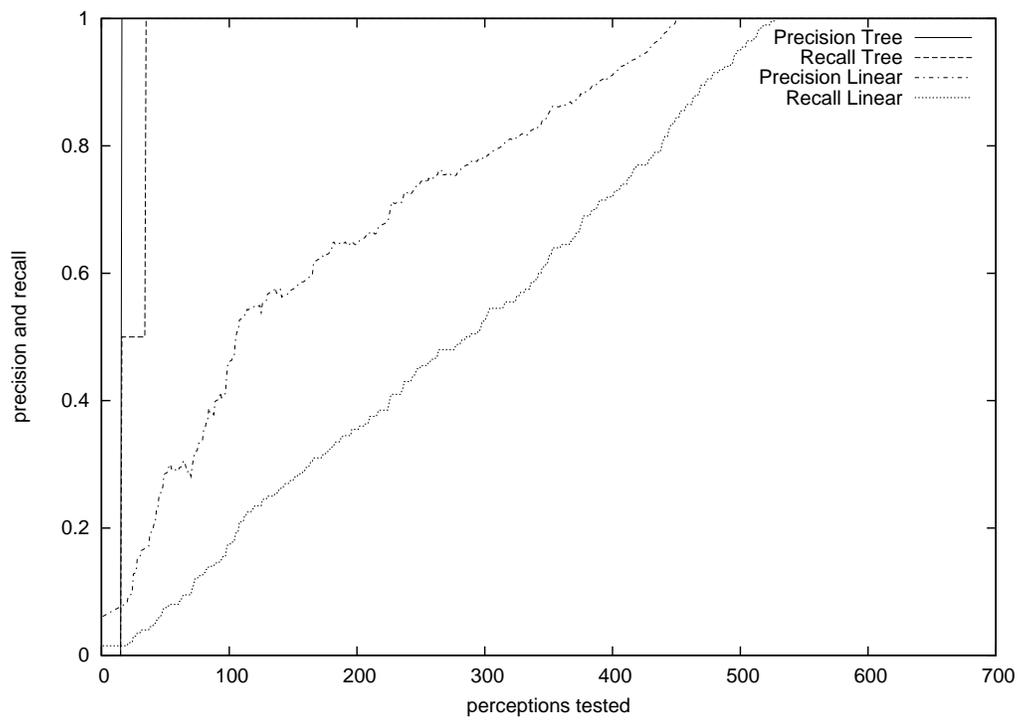}	
\caption{Precision and recall under the number of authorized tests on perceptions for linear Base and tree Base similarity}
\label{fig:courbesProf}
\end{figure}

\newpage

\end{document}